%% file: main.tex
\documentclass[10pt,twocolumn,letterpaper]{article}

\usepackage{cvpr}              
\usepackage{cuted}
\usepackage{pifont}
\usepackage{multirow}
\usepackage{colortbl}
\usepackage{csquotes}


\definecolor{cvprblue}{rgb}{0.21,0.49,0.74}
\usepackage[pagebackref,breaklinks,colorlinks,allcolors=cvprblue]{hyperref}

\def\paperID{*****} 
\def\confName{CVPR}
\def\confYear{2025}

\title{Effective SAM Combination for Open-Vocabulary Semantic Segmentation}

\author{Minhyeok Lee$^{1}$ \quad
	Suhwan Cho$^{1}$ \quad
	Jungho Lee$^{1}$ \quad
	Sunghun Yang$^{1}$ \\
	Heeseung Choi$^{2}$ \quad
	Ig-Jae Kim$^{2}$ \quad
	Sangyoun Lee$^{1}$ \\
	\vspace{-0.1cm}
	$^{1}$Yonsei University\\
	$^{2}$Korea Institute of Science and Technology (KIST)\\
	{\tt\small \{hydragon516, chosuhwan, 2015142131, sunghun98, syleee\}@yonsei.ac.kr}\\
	{\tt\small \{hschoi, drjay\}@kist.re.kr}
}

\begin{document}
	\maketitle
	
	\begin{strip}
		\centering
		\includegraphics[width=1.0\textwidth]{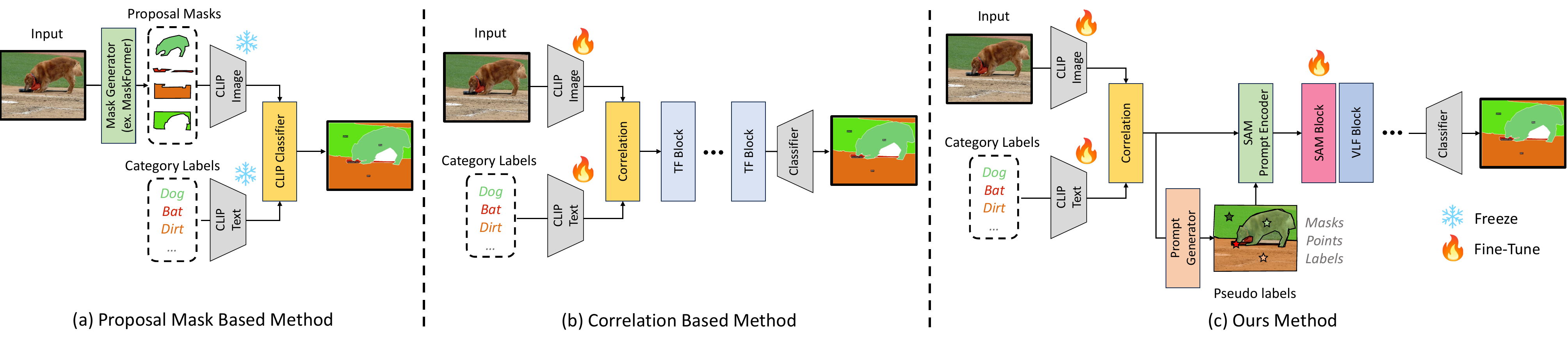}
		\captionof{figure}{
			(a) A model structure that generates proposal masks using a mask generation model. (b) A model structure that refines the correlation between image and text. (c) The structure of the proposed ESC-Net. Our ESC-Net efficiently models the relationship between images and text by combining a pre-trained SAM block with pseudo prompts instead of an inefficient mask generation model. This approach enables much denser mask prediction compared to conventional correlation-based methods.}
		\label{fig:intro}
	\end{strip}
	
	\input{sec/0_abstract}    
	\input{sec/1_intro}
	\input{sec/2_related_works}
	\input{sec/3_approach}

	\input{sec/4_experiments}
	\input{sec/5_conclusion}
	
	\noindent\footnotesize\textbf{Acknowledgement.} This work was supported by the Korea Institute of Science and Technology (KIST) Institutional Program (Project No.2E33001-24-086), National Research Foundation of Korea (NRF) grant funded by the Korea government (MSIT)(No. RS-2024-00423362), and Yonsei Signature Research Cluster Program of 2024 (2024-22-0161).
	{
		\small
		\bibliographystyle{ieeenat_fullname}
		\bibliography{main}
	}
	
	
\end{document}

%% file: sec/0_abstract.tex
\begin{abstract}
Open-vocabulary semantic segmentation aims to assign pixel-level labels to images across an unlimited range of classes. 
Traditional methods address this by sequentially connecting a powerful mask proposal generator, such as the Segment Anything Model (SAM), with a pre-trained vision-language model like CLIP. 
But these two-stage approaches often suffer from high computational costs, memory inefficiencies. 
In this paper, we propose ESC-Net, a novel one-stage open-vocabulary segmentation model that leverages the SAM decoder blocks for class-agnostic segmentation within an efficient inference framework.
By embedding pseudo prompts generated from image-text correlations into SAM’s promptable segmentation framework, ESC-Net achieves refined spatial aggregation for accurate mask predictions. 
Additionally, a Vision-Language Fusion (VLF) module enhances the final mask prediction through image and text guidance. 
ESC-Net achieves superior performance on standard benchmarks, including ADE20K, PASCAL-VOC, and PASCAL-Context, outperforming prior methods in both efficiency and accuracy. 
Comprehensive ablation studies further demonstrate its robustness across challenging conditions.
\end{abstract}

%% file: sec/1_intro.tex
\section{Introduction}
Open-vocabulary semantic segmentation aims to segment an image at the pixel level using an unlimited range of class labels. To achieve this, many approaches have gained attention by leveraging pre-trained vision-language models such as CLIP~\cite{radford2021learning} and ALIGN~\cite{jia2021scaling}, which associate images with various textual descriptions. However, since these models are trained at the image level using paired image-text datasets, they tend to perform poorly on pixel-level tasks.

To address this issue, many methods~\cite{ding2022open, xu2022simple, huynh2022open, liang2023open, wang2024use} adopt a two-stage approach that utilizes a mask proposal generator. As shown in Figure~\ref{fig:intro} (a), these methods first use a powerful class-agnostic mask proposal generator with strong pre-segmentation capabilities, such as MaskFormer~\cite{cheng2021per} or Segment Anything Model (SAM)~\cite{kirillov2023segment}, to pre-segment the image. Next, each image region is fed into a pre-trained CLIP classifier to determine its class. However, these methods have two major issues. First, the two-stage pipeline incurs significant computational costs and inefficient memory usage. Second, because these methods remove the background outside the segmented regions, a domain gap exists between the pre-trained CLIP model and the cropped regions. Additionally, as cropped regions typically include parts of other objects, classification accuracy often suffers.

To overcome these challenges, recent works~\cite{xie2024sed, cho2024cat, jiao2024collaborative} address open-vocabulary semantic segmentation by transferring the modeling of image-text correlation to pixel-level predictions. As shown in Figure~\ref{fig:intro} (b), these methods first establish multi-modal correlations from the embedded CLIP image and text features. Then, by refining this information through the model, they generate the final prediction masks in a manner similar to traditional semantic segmentation. Therefore, unlike the previous two-stage methods, these approaches can perform inference much more efficiently. Additionally, they generally enable direct fine-tuning of CLIP, further optimizing the cross-modality modeling between vision and language. However, CLIP's vision-language contrastive learning aligns only the global representations of context, causing it to focus solely on the text-described information. As a result, these methods perform correlation at a low-resolution, highly embedded feature level to maximize the encoder's localization capability. In other words, while previous two-stage methods can generate high-resolution boundary masks using a mask proposal generator, correlation-based models produce less accurate masks because of CLIP's limited ability to reconstruct detailed local information.

In this paper, we propose a novel one-stage open-vocabulary semantic segmentation model called the Effective SAM Combination (ESC-Net) to address these challenges. Our ESC-Net leverages SAM's powerful class-agnostic segmentation capability while maintaining an efficient inference pipeline. To achieve this, we focus on SAM's ability to effectively transfer to downstream segmentation tasks through promptable segmentation. Figure~\ref{fig:intro} (c) illustrates the proposed ESC-Net's process. First, the proposed ESC-Net computes the correlation between CLIP image features and text features, generating pseudo coordinate points and object masks that are highly correlated with the text class. This pre-generated information is then embedded as prompts via SAM's prompt encoder, and used as pseudo features that incorporate approximate location information. Next, the generated prompt features and CLIP encoder features are used as inputs to the pre-trained SAM transformer blocks, enabling cross-model interaction between the prompts and the image. Therefore, due to SAM’s region-wise spatial aggregation capability, the image features are effectively represented to facilitate modeling the image-language relationships. In other words, this is made possible by SAM’s ability to generate valid segmentation masks even when the prompts are ambiguous and can reference multiple objects. Finally, we design an Vision-Language Fusion (VLF) module to guide the model in generating the final class prediction masks based on image and text guidance.

The ESC-Net structure has two major advantages. First, the combination of correlation-based pseudo prompts and SAM blocks provides more concentrated spatial context information for image-text modeling. This allows for more accurate and dense prediction masks compared to previous one-stage methods that relied on CLIP’s weak spatial localization signals for images and text. Second, since ESC-Net does not use the SAM image encoder, it offers much more efficient inference compared to the existing two-stage methods that utilize the entire SAM pipeline. Notably, the proposed method achieves better performance compared to the previous correlation-based one-stage method, CAT-Seg~\cite{cho2024cat}, despite having similar computational costs.

The proposed ESC-Net is evaluated on standard open-vocabulary benchmarks, including the ADE20K~\cite{zhou2019semantic}, PASCAL-VOC~\cite{everingham2010pascal}, and PASCAL-Context~\cite{mottaghi2014role} datasets, achieving state-of-the-art performance compared to previous methods. Additionally, we demonstrate through various ablation studies that our model efficiently maintains strong performance across a range of challenging scenarios.

Our main contributions can be summarized as follows:

\begin{itemize}
	\item We propose ESC-Net, a novel one-stage open-vocabulary semantic segmentation model. It effectively combines CLIP and SAM to leverage SAM's powerful class-agnostic segmentation capabilities while maintaining efficient inference.
	
	\item ESC-Net introduces correlation-based pseudo prompts derived from CLIP image-text features. These prompts are embedded through SAM's prompt encoder and used to guide the SAM transformer blocks. This enhances image-text modeling and results in more accurate, dense prediction masks.
	
	\item Our method achieves state-of-the-art performance on standard open-vocabulary benchmarks such as ADE20K, PASCAL-VOC, and PASCAL-Context. It outperforms previous one-stage and two-stage methods with similar computational costs. Extensive ablation studies demonstrate its robust performance across various challenging scenarios.
\end{itemize}

%% file: sec/2_related_works.tex
\section{Related Work}
\textbf{Open-vocabulary semantic segmentation.} Open vocabulary semantic segmentation evolves as a specialized task within computer vision, aiming to enable models to segment arbitrary, unseen categories beyond predefined labels. Early methods~\cite{bucher2019zero, xian2019semantic, zhao2017open} in this area focus on aligning visual features with pre-trained text embeddings by learning a feature mapping that associates visual and text spaces effectively.

\begin{figure*}[t]
	\setlength{\belowcaptionskip}{-24pt}
	\begin{center}
		\includegraphics[width=\linewidth]{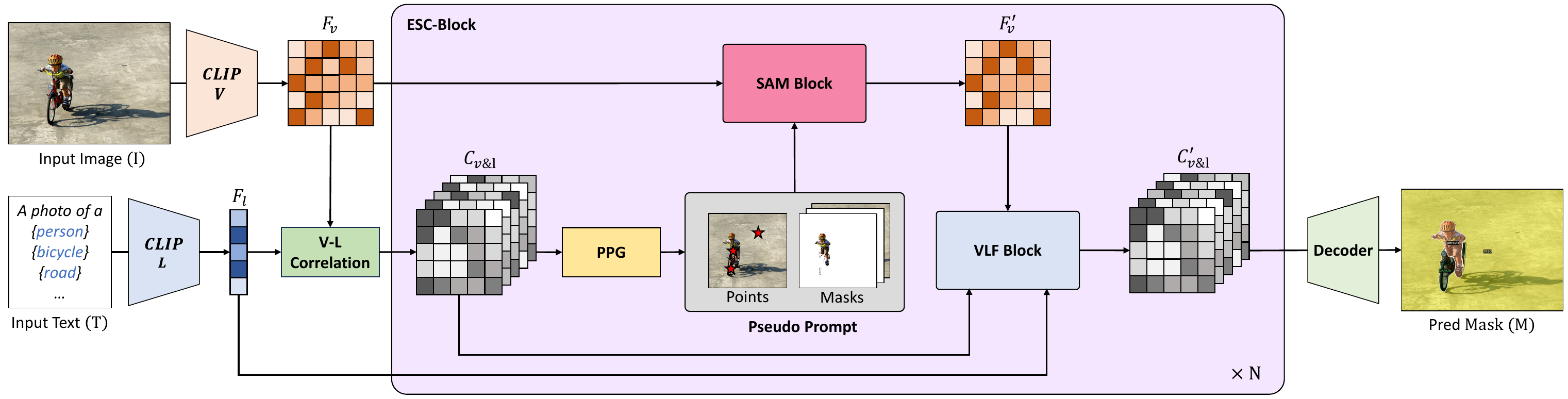}
		\caption{The proposed ESC-Net consists of the CLIP vision and language encoders, $N$ consecutive ESCBlocks, and a decoder. Each ESC-Block generates a pseudo prompt from the image-text correlation map and uses it as input to the SAM block. The SAM block aggregates the CLIP image features. The VLF block models the image-text correlation using image features and text features, refining the correlation map through this process.}
		\label{fig:main}
	\end{center}
\end{figure*}

With the advent of vision-language models like CLIP, many works leverage its extensive vocabulary and visual-text alignment capabilities to address the vocabulary limitations of previous approaches. For instance, two-stage frameworks like OpenSeg~\cite{ghiasi2022scaling}, ZegFormer~\cite{ding2022decoupling}, and ZSseg~\cite{xu2022simple} use class-agnostic mask proposals that are later matched with CLIP-based text embeddings for classification. OVSeg~\cite{liang2023open} improves this two-stage approach by fine-tuning CLIP with region-text pairs to enhance segmentation accuracy on diverse categories, while ODISE~\cite{xu2023open} employs a diffusion model to generate high-quality masks, thus bypassing the reliance on pre-trained mask generators. However, these methods often depend on region generators, which are typically trained on limited annotated data, thus constraining generalization to unseen categories.

Single-stage frameworks emerge as an alternative to bypass the need for region proposals. Models such as LSeg~\cite{li2022language} and MaskCLIP~\cite{ding2022open} adopt pixel-level learning directly from CLIP’s text embeddings, aiming to predict segmentation maps in one forward pass. Methods like SAN~\cite{xu2023side} and ZegCLIP~\cite{zhou2023zegclip} introduce side adapter networks and learnable tokens to assist CLIP's segmentation performance. CAT-Seg~\cite{cho2024cat} adapts CLIP by aggregating cosine similarity between image and text embeddings, fine-tuning the encoders to enhance segmentation on unseen classes. Meanwhile, MAFT+~\cite{jiao2024collaborative} proposes a collaborative optimization approach, where content-dependent transfer adaptively refines text embeddings based on image context, while a Representation Compensation strategy preserves CLIP’s zero-shot capability.

\noindent
\textbf{Segment Anything Model.}
The Segment Anything Model (SAM)\cite{kirillov2023segment}, recently introduced by Meta AI Research, marks a pivotal advancement in the field of foundation models for computer vision, particularly image segmentation. It builds on the Vision Transformer\cite{dosovitskiy2020image} architecture and trains on the vast SA-1B dataset, which contains over a billion segmentation masks across 11 million images. SAM positions itself as a transformative tool in zero-shot image segmentation. Its primary innovation lies in its “promptable” design, which enables it to generalize across various image domains without fine-tuning, allowing effective segmentation in new environments through prompt engineering. This promptability aligns SAM with trends in natural language processing, where large language models, also pre-trained on web-scale data, demonstrate robust zero-shot and few-shot capabilities.

Prior to SAM, foundation models in computer vision, such as CLIP and ALIGN, focus on aligning multimodal data—specifically text and image encoders—through contrastive learning for cross-modal zero-shot generalization. However, these earlier models primarily excel at tasks reliant on cross-modal alignment, such as image captioning and generation, rather than segmentation. SAM advances beyond these models by specifically targeting image segmentation as a foundational task, making it a versatile tool for addressing complex segmentation challenges.

%% file: sec/3_approach.tex
\section{Proposed Approach}
\subsection{Overall Architecture}

Figure~\ref{fig:main} shows the overall structure of the proposed ESC-Net. Our model consists of a CLIP vision encoder, language encoder, and a single decoder for mask prediction, similar to conventional one-stage methods, without the need for an additional mask proposal model. In this task, we provide the input image $I$ and a set of text descriptions $T$ for all candidate classes to the model, where $T^n$ denotes the text description of the $n$-th class. The total number of candidate classes is $N_c$, where $n=$ $1, \ldots, N_c$. Therefore, ESC-Net has the same input configuration as conventional one-stage methods. First, ESC-Net generates a vision-language correlation map $C_{v\&l}$ from the CLIP vision feature $F_v \in \mathbb{R} ^ {C \times H \times W}$ and language feature $F_l \in \mathbb{R} ^ {C \times N_c}$. $C_{v\&l} \in \mathbb{R} ^ {N_c \times H \times W}$ represents the image similarity relationships for each text class. Next, we design a Pseudo Prompt Generator (PPG) to generate pseudo point coordinates and mask regions for each class candidate based on $C_{v\&l}$. These pseudo prompts embed approximate location information about the object regions and are passed to the pre-trained SAM block. A detailed explanation of the PPG is provided in Section~\ref{sec:ppg}. Next, $F_v$ passes through the SAM transformer block along with the pseudo prompts, refining it into $F'_v \in \mathbb{R} ^ {C \times H \times W}$ with enhanced class-agnostic spatial context information. Finally, we propose the Vision-Language Fusion (VLF) module to model the interaction between the refined vision feature and the text feature. In this module, $C_{v\&l}$ is aggregated with $F_v$ and $F_l$ to produce a reconstructed correlation map $C'_{v\&l} \in \mathbb{R} ^ {N_c \times C \times H \times W}$. Details on the VLF module are discussed in Section~\ref{sec:vlf}. As shown in Figure~\ref{fig:main}, we progressively refine $C_{v\&l}$ using $N$ ESC blocks from the above process and generate the final prediction mask $M$ through the mask decoder.

\begin{figure}[t]
	\setlength{\belowcaptionskip}{-24pt}
	\begin{center}
		\includegraphics[width=\linewidth]{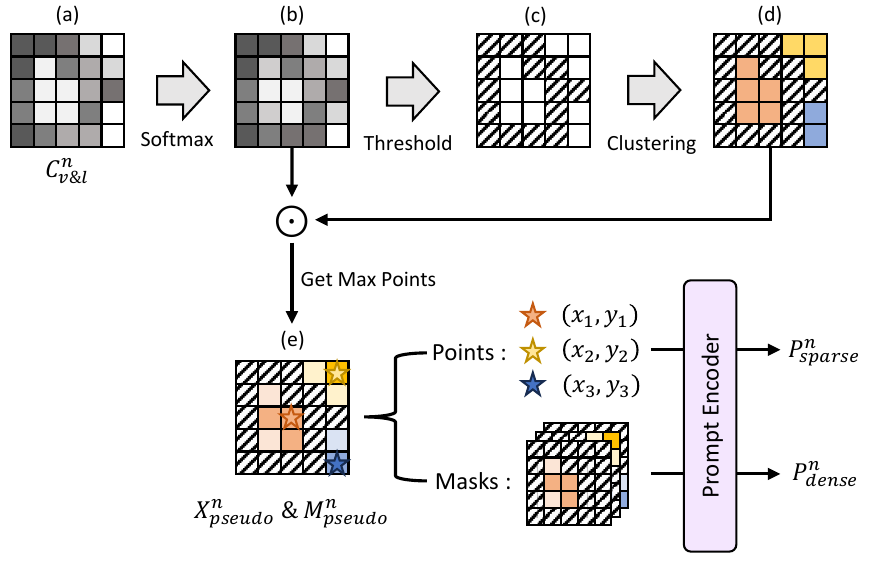}
		\caption{The process of the proposed Pseudo Prompt Generator (PPG). PPG aims to generate class-specific pseudo prompts from image-text correlation maps. For efficiency, all processes are computed in batch-wise parallelization over all classes.}
		\label{fig:ppg}
	\end{center}
\end{figure}

\subsection{Vision-Language Correlation}
Before generating the pseudo prompt, we first create a correlation map using the correlation between $F_v$ and $F_l$. This process is performed based on simple cosine similarity. Let $C^n_{v\&l}\left(i\right)$ be the image correlation map for the $n$-th class information, and let $i$ represent the 2D spatial coordinates of the image embedding. Then, $C^n_{v\&l}\left(i\right)$ is expressed as follows:

\begin{equation}
	C_{v\&l}^n(i)=\frac{F_v(i) \cdot F_l^n}{\left\|F_v(i)\right\|\left\|F_l^n\right\|}.
\end{equation}

\subsection{Pseudo Prompt Generator}
\label{sec:ppg}
The purpose PPG aim to generate location information for candidate objects from $C^n_{v\&l}$ and convert it into the SAM prompt format. 
Traditional SAM~\cite{kirillov2023segment} can take some or all of points, bounding boxes, masks, and text as input prompts. Among these, we use only points and masks as prompts. Figure~\ref{fig:ppg} shows the proposed PPG process. 

First, PPG applies a softmax operation to the correlation map $C^n_{v\&l} \in \mathbb{R} ^ {1 \times H \times W}$ for each $n$-th class to generate class-specific object probability masks (b). Then, it binarizes this probability mask using a predefined threshold value $\alpha$. As a result, this binary mask (c) represents the approximate location of candidate objects corresponding to class $n$. However, since multiple objects may exist for a single class, we cluster the mask by region, as shown in Figure~\ref{fig:ppg} (d). To achieve this, we apply a k-means clustering algorithm~\cite{hartigan1979algorithm} based on pixel locations to divide the binary mask into $N_o$ mask regions. Next, we multiply the probabilistic mask (b) by the clustered mask region map (d) to generate a probability map (e) where only the object regions are filtered. Finally, we generate pseudo points and pseudo masks from (e). As shown in Figure 3, the pseudo points are determined as the highest probability areas for each object region in (e), and the pseudo masks correspond to the masks of each region. Consequently, a total of $N_o$ pseudo points and $N_o$ pseudo masks are generated.

The generated pseudo points and pseudo masks are used as inputs to the prompt encoder, as proposed by the traditional SAM. According to SAM's approach, the points and masks are embedded as sparse prompt features $P^n_{sparse}$ and dense prompt features $P^n_{dense}$, respectively.

\subsection{SAM Block}
\label{sec:sam_block}

\begin{figure}[t]
	\setlength{\belowcaptionskip}{-24pt}
	\begin{center}
		\includegraphics[width=\linewidth]{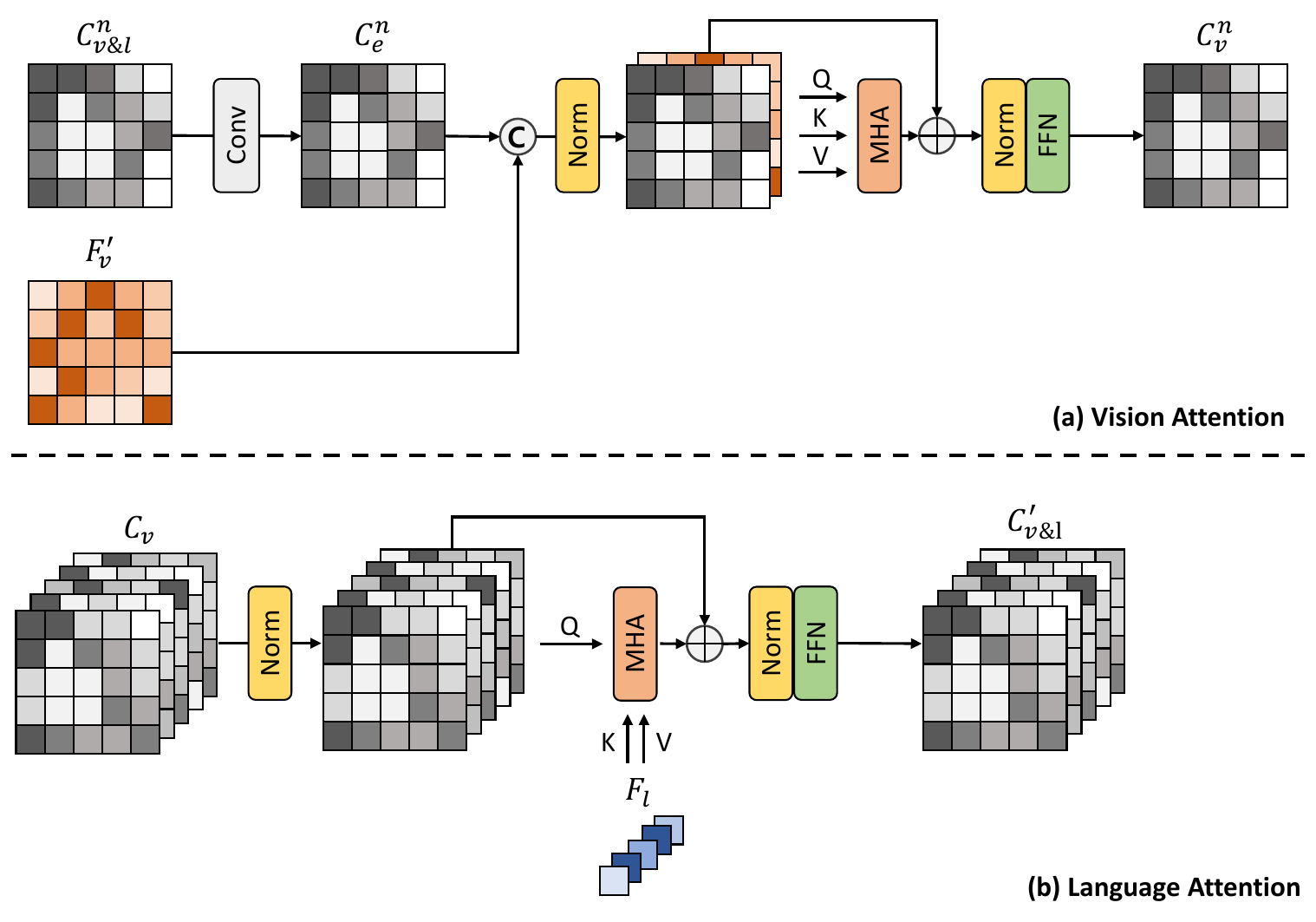}
		\caption{The structure of the proposed Vision-Language Fusion (VLF) block. VLF sequentially applies image and text guidance to the correlation map to refine it.}
		\label{fig:vlf}
	\end{center}
\end{figure}

The proposed ESC-Net utilizes a pre-trained SAM transformer block, as shown in Figure~\ref{fig:main}, to aggregate the spatial context information of image features using generated pseudo-prompt information. Specifically, we leverage the transformer decoder block of the SAM mask decoder, which effectively maps images and prompts. The SAM block first applies a prompt self-attention layer $SA\left(.\right)$, followed by bidirectional cross-attention $BCA\left(.,.\right)$ for image-to-prompt and prompt-to-image, updating the image embedding feature $F_v$ to $F_v^{\prime} \in \mathbb{R} ^ {C \times H \times W}$. However, unlike the original SAM, we use $N_c$ sets of prompts, so the operations for each prompt are parallelized in batches. Therefore, the refined image feature $\left(F^n_v\right)^{\prime} \in \mathbb{R} ^ {C \times H \times W}$ for the $n$-th prompt set is expressed as follows:

\begin{equation}
	\begin{aligned}
			\left(F_v^n\right)^{\prime}=B C A\left(S A\left(F_l^n\right), F_v\right), \\
			F_v^{\prime}=\operatorname{Conv}\left(\left[F_v^0 ; F_v^1 ; \ldots ; F_v^{N_c}\right]\right),
	\end{aligned}
\end{equation}
\noindent
where $Conv(.)$ denotes a $1 \times 1$ convolution, and $[.;.]$ represents the concatenation operation. In this paper, we apply a total of $N$ ESC blocks, thus utilizing the parameters of $N$ consecutive pre-trained SAM blocks.

\begin{table*}[!t]
	\begin{center}
		\resizebox{\textwidth}{!}{
			\begin{tabular}{l|ccccc|cccccc}
				\toprule
				Model & Publication & VLM & Additional Backbone & Training Dataset & Additional Dataset & A-847 & PC-459 & A-150 & PC-59 & PAS-20 & $\textnormal{PAS-20}^b$\\
				\midrule\midrule
				SPNet~\citep{xian2019semantic} & CVPR'19 & - & ResNet-101 & PASCAL VOC & \ding{55} & - & - & - & 24.3 & 18.3 & - \\
				ZS3Net~\citep{bucher2019zero}  & NeurIPS'19 & - & ResNet-101 & PASCAL VOC & \ding{55}  & - & - & - & 19.4 & 38.3 & - \\
				LSeg~\citep{li2022language} & ICLR'22 & CLIP ViT-B/32 & ResNet-101 & PASCAL VOC-15  & \ding{55} & - & - & - & - & 47.4 & - \\
				LSeg+~\citep{ghiasi2022scaling} & ECCV'22 & ALIGN & ResNet-101 & COCO-Stuff & \ding{55}  & 2.5 & 5.2 & 13.0 & 36.0 & - & 59.0 \\
				ZegFormer~\citep{ding2022decoupling} & CVPR'22 & CLIP ViT-B/16 & ResNet-101 & COCO-Stuff-156 & \ding{55} & 4.9 & 9.1 & 16.9 & 42.8 & 86.2 &  62.7 \\
				ZSseg~\citep{xu2022simple} & ECCV'22 & CLIP ViT-B/16 & ResNet-101 & COCO-Stuff & \ding{55}   & 7.0 & - & 20.5 & 47.7 & 88.4 & - \\
				OpenSeg~\citep{ghiasi2022scaling} & ECCV'22 & ALIGN & ResNet-101 & COCO Panoptic & \ding{51}  &  4.4 & 7.9 & 17.5 & 40.1 & - & 63.8 \\
				OVSeg~\citep{liang2023open} & CVPR'23 & CLIP ViT-B/16 & ResNet-101c & COCO-Stuff & \ding{51}  &  7.1 & 11.0 & 24.8 & 53.3 & 92.6 & - \\
				ZegCLIP~\citep{zhou2023zegclip} & CVPR'23 & CLIP ViT-B/16 & - & COCO-Stuff-156 & \ding{55}  & -&-&-&41.2& 93.6 & - \\
				SAN~\citep{xu2023side} & CVPR'23 & CLIP ViT-B/16 & - & COCO-Stuff & \ding{55}  & 10.1 & 12.6 & 27.5 & 53.8 & 94.0 & - \\
				DeOP~\citep{han2023open} & ICCV'23 & CLIP ViT-B/16 & ResNet-101c & COCO-Stuff-156 & \ding{55}  & 7.1 & 9.4 & 22.9 &  48.8 & 91.7 & - \\
				SCAN~\citep{liu2024open} & CVPR'24 & CLIP ViT-B/16 & Swin-B & COCO-Stuff & \ding{55}  & 10.8 & 13.2 & 30.8 & \underline{58.4} & \underline{97.0} & - \\
				EBSeg~\citep{shan2024open} & CVPR'24 & CLIP ViT-B/16 & SAM ViT-B & COCO-Stuff & \ding{55}  & 11.1 & 17.3 & 30.0 & 56.7 & 94.6 & - \\
				SED~\citep{xie2024sed} & CVPR'24 & ConvNeXt-B & - & COCO-Stuff & \ding{55}  & 11.4 & 18.6 & 31.6 & 57.3 & 94.4 & - \\
				CAT-Seg~\citep{cho2024cat} & CVPR'24 & CLIP ViT-B/16 & - & COCO-Stuff & \ding{55}  & \underline{12.0} & \underline{19.0} & \underline{31.8} & 57.5 & 94.6 & \underline{77.3} \\
				\rowcolor{black!6} 
				& & & & & &  \textbf{13.3} &\textbf{21.1} &\textbf{35.6} &\textbf{59.0} & \textbf{97.3} & \textbf{80.1} \\
				\rowcolor{black!6}\multirow{-2}{*}{ESC-Net (ours)} & \multirow{-2}{*}{-} & \multirow{-2}{*}{CLIP ViT-B/16} & \multirow{-2}{*}{-} & \multirow{-2}{*}{COCO-Stuff} & \multirow{-2}{*}{\ding{55}}  & \textcolor{VioletRed}{(+1.3)} & \textcolor{VioletRed}{(+2.1)} & \textcolor{VioletRed}{(+3.8)} & \textcolor{VioletRed}{(+0.6)} & \textcolor{VioletRed}{(+0.3)} & \textcolor{VioletRed}{(+2.8)} \\
				\midrule
				LSeg~\citep{li2022language} & ICLR'22 & CLIP ViT-B/32 & ViT-L/16 & PASCAL VOC-15 & \ding{55} & - & - & - & - & 52.3 & - \\
				OpenSeg~\citep{ghiasi2022scaling} & ECCV'22 & ALIGN & EfficientNet-B7 & COCO Panoptic & \ding{51} & 8.1 & 11.5 & 26.4 & 44.8 & - & 70.2\\
				OVSeg~\citep{liang2023open} & CVPR'23 & CLIP ViT-L/14 & Swin-B & COCO-Stuff & \ding{51} & 9.0 & 12.4 & 29.6 & 55.7 & 94.5 & - \\
				SAN~\citep{xu2023side} & CVPR'23 & CLIP ViT-L/14 & - & COCO-Stuff & \ding{55} & 12.4 & 15.7 & 32.1 & 57.7 & 94.6 & - \\
				ODISE~\citep{xu2023open} & CVPR'23 & CLIP ViT-L/14 & Stable Diffusion & COCO-Stuff & \ding{55} & 11.1 & 14.5 & 29.9 & 57.3 & - & - \\
				FC-CLIP~\citep{yu2023convolutions} & NeurIPS'23 & ConvNeXt-L & - & COCO Panoptic  & \ding{55} & 11.2 & 12.7 & 26.6 & 42.4 & 89.5 & - \\
				MAFT~\citep{jiao2023learning} & NeurIPS'23 & CLIP ViT-L/14 & - & COCO-Stuff & \ding{55} & 12.7 & 16.2 & 33.0 & 59.0 & 92.1 & - \\
				USE~\citep{wang2024use} & CVPR'24 & CLIP ViT-L/14 & DINOv2, SAM & COCO-Stuff & \ding{51} & 13.4 & 15.0 & 37.1 & 58.0 & - & - \\
				SCAN~\citep{liu2024open} & CVPR'24 & CLIP ViT-L/14 & Swin-B & COCO-Stuff & \ding{55}  & 14.0 & 16.7 & 33.5 & 59.3 & 97.0 & - \\
				EBSeg~\citep{shan2024open} & CVPR'24 & CLIP ViT-L/14 & SAM ViT-B & COCO-Stuff & \ding{55} & 13.7 & 21.0 & 32.8 & 60.2 & \underline{97.2} & - \\
				SED~\citep{xie2024sed} & CVPR'24 & ConvNeXt-L & - & COCO-Stuff & \ding{55} & 13.9 & 22.6 & 35.2 & 60.6 & 96.1 & - \\
				CAT-Seg~\citep{cho2024cat} & CVPR'24 & CLIP ViT-L/14 & - & COCO-Stuff & \ding{55} & \underline{16.0} & \underline{23.8} & \underline{37.9} & \underline{63.3} & 97.0 & \underline{82.5} \\
				MAFT+~\citep{jiao2025collaborative} & ECCV'24 & ConvNeXt-L & - & COCO-Stuff & \ding{55} & 15.1 & 21.6 & 36.1 & 59.4 & 96.5 & - \\
				\rowcolor{black!6} 
				& & & & & & \textbf{18.1} & \textbf{27.0} & \textbf{41.8} & \textbf{65.6} & \textbf{98.3} & \textbf{86.3}\\
				\rowcolor{black!6}\multirow{-2}{*}{ESC-Net (ours)} & \multirow{-2}{*}{-} & \multirow{-2}{*}{CLIP ViT-L/14} & \multirow{-2}{*}{-} & \multirow{-2}{*}{COCO-Stuff} & \multirow{-2}{*}{\ding{55}} & \textcolor{VioletRed}{(+2.1)} & \textcolor{VioletRed}{(+3.2)} & \textcolor{VioletRed}{(+3.9)} & \textcolor{VioletRed}{(+2.3)} & \textcolor{VioletRed}{(+1.1)} & \textcolor{VioletRed}{(+3.8)}\\
				
				\bottomrule
			\end{tabular}
		}
		\caption{Quantitative evaluation on open-vocabulary segmentation benchmarks. The best-performing results are presented in bold, while the second-best results are underlined. Improvements over the second-best are highlighted in \textcolor{VioletRed}{red}.}
		\label{tab:main_table}
	\end{center}
\end{table*}

\subsection{Vision-Language Fusion Module}
\label{sec:vlf}

The VLF module aims to refine the correlation map $C_{v\&l}$ by using the image embedding feature $F'_v$, which has enhanced spatial aggregation through SAM blocks, and the text feature $F_l$. First, as shown in Figure~\ref{fig:vlf} (a), the correlation map $C^n_{v\&l} \in \mathbb{R} ^ {1 \times H \times W}$ for the $n$-th class is embedded using a $1 \times 1$ convolution layer. The embedded class correlation map $C^n_e \in \mathbb{R} ^ {C \times H \times W}$ is then concatenated with $F'_v$ and used as input to the transformer block. To improve the model's efficiency, this process is performed with batch-level parallel operations for all classes, similar to the SAM block. Furthermore, similar to CAT-Seg~\cite{cho2024cat}, we structure the transformer block using the Swin Transformer~\cite{liu2021swin} architecture to gain additional memory advantages. As a result of the above process, the vision correlation map $C^n_v \in \mathbb{R} ^ {C \times H \times W}$ for the $n$-th class is generated, as shown in Figure~\ref{fig:vlf} (a). Next, as shown in Figure~\ref{fig:vlf} (b), we introduce an attention mechanism with $F_l$ as the key and value to explicitly capture relationships between different class categories. For open-vocabulary segmentation, the number of input classes $N_c$ can vary, and it needs to be order-agnostic. Therefore, we adopt a linear transformer~\cite{katharopoulos2020transformers} structure without positional embedding. As a result, the VLF module generates a multimodally refined embedding correlation map $C'_{v\&l}$.

\subsection{Mask Prediction Decoder}
\label{sec:decoder}
We use the simple upsampling layers of a U-Net~\cite{ronneberger2015u} architecture, composed of bilinear interpolation and $3 \times 3$ convolution layers, to generate the final predicted segmentation map. This approach is similar to existing one-stage methods and employs CLIP image encoder features as skip connections.

%% file: sec/4_experiments.tex
\section{Experiments}
\subsection{Datasets}
In this paper, we train the proposed ESC-Net on the COCO-Stuff~\cite{caesar2018coco} dataset, following established methods. The COCO-Stuff training set includes approximately 118k images annotated with masks across 171 semantic categories. To ensure a fair comparison with previous methods, we evaluate our model on the ADE20K~\cite{zhou2019semantic}, PASCAL-VOC~\cite{everingham2010pascal}, and PASCAL-Context~\cite{mottaghi2014role} datasets. Details for each dataset are as follows:

\begin{figure*}[t]
	\centering
	\includegraphics[width=\linewidth]{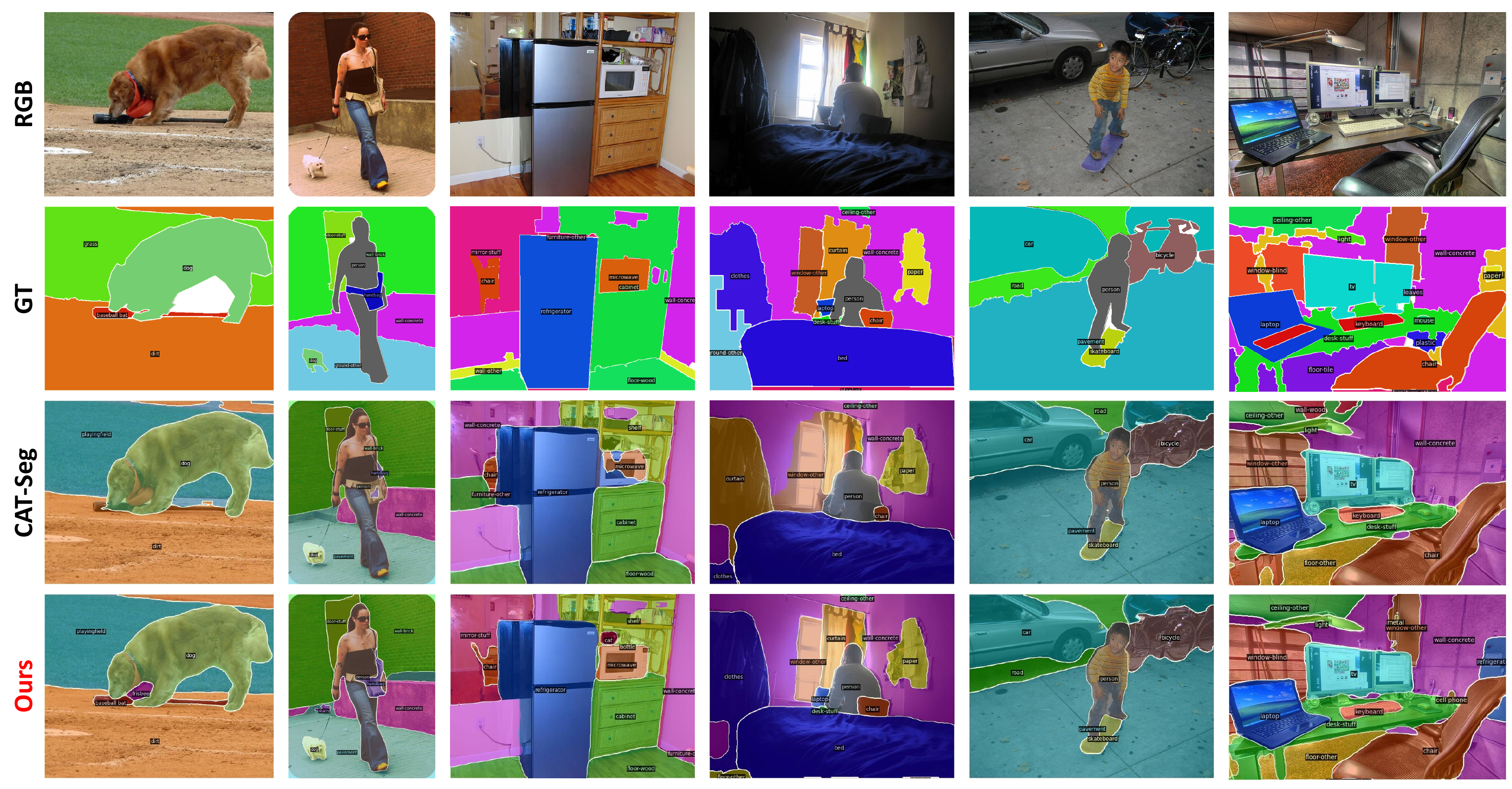}
	\caption{Qualitative comparison of CAT-Seg and our ESC-Net across various datasets. Our model is capable of generating more accurate and robust masks compared to existing correlation-based state-of-the-art method.}
	\label{fig:result}
\end{figure*}

\noindent
\textbf{ADE20K}~\cite{zhou2019semantic} is a large-scale semantic segmentation dataset featuring 20,000 training images and 2,000 validation images. In open-vocabulary semantic segmentation, it includes two distinct test sets: A-150, containing 150 commonly seen categories, and A-847, which includes a total of 847 categories.

\noindent
\textbf{PASCAL-VOC}~\cite{everingham2010pascal}, one of the foundational datasets for object detection and segmentation, has approximately 1,500 training images and 1,500 validation images, covering 20 object categories. For the open-vocabulary segmentation task, this dataset is referred to as PAS-20. Additionally, following the approaches in~\cite{ghiasi2022scaling, cho2024cat}, we include PAS-20$^b$ in the evaluation dataset. Unlike PC-59, PAS-20$^b$ is defined without including the “background” class.

\noindent
\textbf{PASCAL-Context}~\cite{mottaghi2014role}, an extension of PASCAL VOC for semantic segmentation, also features two test sets for open-vocabulary segmentation: PC-59 with 59 categories and PC-459 with 459 categories.

\subsection{Evaluation Metric}
To quantitatively evaluate the performance, we follow existing traditional open-vocabulary semantic segmentation. Semantic segmentation results are evaluated with mean Intersection over Union (mIoU)~\cite{everingham2010pascal}.

\subsection{Implementation Details}
In this study, we set the class-specific object count $N_0$ in PPG to 5 and the number of ESC-Blocks $N$ to 4. In other words, PPG generates 5 points and masks per class, and the correlation map is refined through 4 ESC-Blocks after the CLIP encoder. Therefore, we take 4 SAM mask decoder transformer blocks from the pretrained SAM model. The input image is resized to $336 \times 336$, and the AdamW~\cite{loshchilov2017decoupled} optimizer is used for training. For optimal training, the CLIP encoder and SAM blocks are trained with a learning rate of $2 \times 10^{-6}$, while the rest of the model part is trained with a learning rate of $2 \times 10^{-4}$. All experiments are conducted using the PyTorch~\cite{paszke2019pytorch} framework, and training is performed on four NVIDIA RTX A6000 GPUs.

In traditional open-vocabulary segmentation methods~\cite{ghiasi2022scaling, ding2022decoupling, xu2022simple}, the image-text relationship modeling of pre-trained CLIP is typically preserved by freezing its encoders. However, recent studies have demonstrated that fine-tuning CLIP's image and text encoders can achieve higher performance in open-vocabulary segmentation tasks. This suggests that CLIP’s linguistic discrimination capability can be retained even with training datasets containing a limited number of classes. Following this insight, the proposed ESC-Net adopts the approach of CAT-Seg~\cite{cho2024cat} to fine-tune the attention layers of CLIP’s image and text encoders.

\begin{figure}[t]
	\centering
	\includegraphics[width=\linewidth]{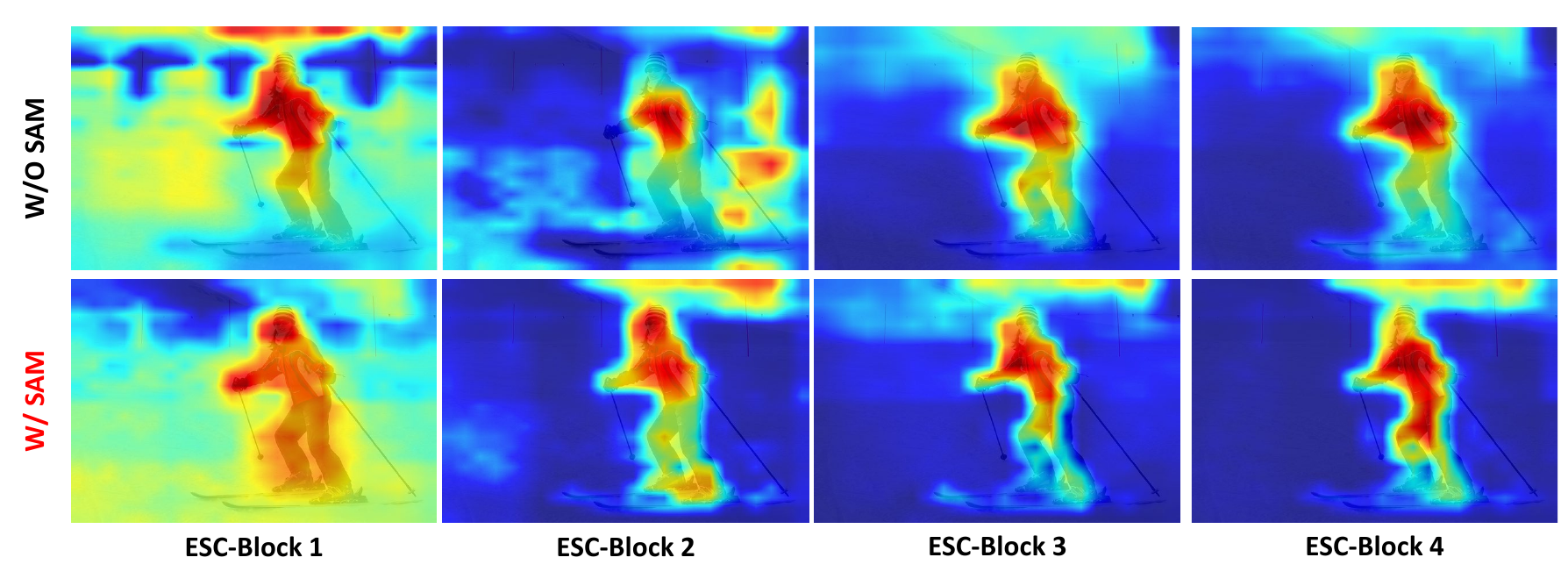}
	\caption{Visualization of image-text correlation maps with and without the SAM block. We visualize the model activation maps for the \enquote{Person} class for each ESC-Block. The proposed SAM-based method enables more accurate and dense object localization compared to the baseline.}
	\label{fig:sam}
\end{figure}

\subsection{Results}
\noindent
\textbf{Quantitative results.} Table~\ref{tab:main_table} presents the quantitative results of the proposed ESC-Net compared to previous state-of-the-art methods on standard open-vocabulary semantic segmentation benchmark datasets~\cite{zhou2019semantic, everingham2010pascal, mottaghi2014role}. In open-vocabulary semantic segmentation tasks, the final performance generally varies depending on the type of vision-language model (VLM) used. Therefore, as shown in Table~\ref{tab:main_table}, we perform comparisons between models with similar-scale VLMs. Some approaches utilize auxiliary models such as SAM or mask formers for mask generation or additional information extraction, in addition to the VLM. Furthermore, some models enhance performance by leveraging additional training datasets~\cite{chen2015microsoft, pont2020connecting}. Hence, in Table~\ref{tab:main_table}, we provide information on the additional backbones and types of training datasets used for each method. As shown in Table 1, the proposed ESC-Net achieves state-of-the-art performance, surpassing existing methods with both CLIP ViT-B/16 and CLIP ViT-L/14 VLMs. Compared to previous state-of-the-art methods, MAFT+~\cite{jiao2025collaborative} and CAT-Seg~\cite{cho2024cat}, there is a significant performance improvement across all evaluation datasets. Notably, ESC-Net shows an mIoU improvement of 2.1 on the A-847 dataset, which includes a large number of classes, and 3.2 on the PC-459 dataset. Furthermore, in Section~\ref{sec:ablation}, we demonstrate the effectiveness of the proposed ESC-Net over existing models through various ablation studies.

\noindent
\textbf{Qualitative results.} Figure~\ref{fig:result} presents the qualitative results comparing the proposed ESC-Net with the previous state-of-the-art single-stage model, CAT-Seg~\cite{cho2024cat}, across various benchmarks. As shown, our model demonstrates robustness to variations in object scale, lighting, and more, achieving denser segmentation compared to CAT-Seg. Furthermore, in scenarios with a high number of objects and complex backgrounds, the proposed model exhibits superior performance. This is due to the effective use of features from the SAM block, which learns class-agnostic segmentation across various scales, for mask generation.

\subsection{Ablation Analysis.}
\label{sec:ablation}
This section includes various ablation experiments on the proposed model. All experiments are evaluated using the CLIP ViT-L/14 VLM with an image size of $336 \times 336$.

\noindent
\textbf{Effect of SAM blocks.} Table~\ref{tab:sam} demonstrates the effectiveness of the pre-trained SAM blocks in the proposed model. First, (a) represents the model without using the SAM blocks, meaning that $F_v$ is not refined. Consequently, the proposed PPG is also removed in this experiment. Next, (b) shows the model that utilizes the SAM structure but does not use pre-trained parameters. Lastly, (c) corresponds to the model that uses the pretrained SAM blocks, which is identical to the proposed ESC-Net. As shown in the table, instead of generating masks using a mask proposal network, significant performance improvements can be achieved by integrating pretrained SAM transformer blocks. In particular, when comparing (b) and (c), which have the same computational load, the approach using pretrained SAM blocks reaches much better segmentation performance. This demonstrates that the proposed ESC-Net effectively utilizes the spatial aggregation capabilities learned by SAM through the pseudo prompts.

Furthermore, to verify that the proposed SAM integration structure aids in image-text modeling, we visualize $C'_{v \& l}$ with and without the SAM block as class-specific activation maps, as shown in Figure 6. Since we include one SAM block in each ESC block, we generate a total of four activation maps. As illustrated in the figure, the proposed SAM integration method shows more accurate and denser mapping to the class target objects compared to the baseline. This demonstrates that the ESC-Net structure can achieve more precise image-text modeling compared to existing methods.

\noindent
\textbf{Effect of the Pseudo Prompt Generation Method.} Table~\ref{tab:ppg} presents the performance of the proposed model based on different combinations of pseudo prompt generation methods. Since the traditional SAM model takes points and bounding boxes as sparse prompt embeddings and masks as dense prompt embeddings, we explore various combinations of prompt embeddings. In this experiment, we utilize the clustered masks shown in Figure~\ref{fig:ppg} (d) to generate bounding box pseudo prompts. For each mask, we find the smallest bounding box that covers it and use this as the pseudo bounding box. Similar to points and masks, 5 bounding boxes per class are sampled. As shown in Table~\ref{tab:ppg} (d), the best performance was achieved when using both points and masks together. In contrast, bounding boxes, as in (a) and (b), were found to be less effective as pseudo prompt embeddings than points. This is likely because the accuracy of the generated pseudo bounding boxes is relatively low, and multiple objects can be overlapped within the box. Although the bounding boxes are generated from pseudo masks, in traditional SAM, mask prompts are trained as areas that approximately indicate the object's location. Therefore, they function effectively as pseudo prompts that are less dependent on the precise position of the object. Consequently, using all three pseudo prompts, as in Table~\ref{tab:ppg} (e), is inefficient; thus, we use only points and masks as pseudo prompts.

\begin{table}[t]
	\centering
	\resizebox{0.48\textwidth}{!}{%
		\begin{tabular}{cl|cccccc}
			\toprule
			Index & Methods            & A-847 & PC-459 & A-150 & PC-59 & PAS-20 & PAS-20 \\
			\midrule\midrule
			(a)   & \textit{W/O} SAM  & 4.8 & 11.7 & 24.2 & 50.4 & 89.4 & 74.8 \\
			(b)   & \textit{W/} \,\; SAM~(\textit{rand init})   & 5.9 & 15.8 & 28.4 & 55.9 & 91.5 & 77.2 \\
			(c)   & \textit{W/} \,\; SAM~(\textit{pretrained}) & \textbf{18.1} & \textbf{27.0} & \textbf{41.8} & \textbf{65.6} & \textbf{98.3} & \textbf{86.3} \\
			\bottomrule
		\end{tabular}
	}
	\caption{Quantitative performance comparison based on the use of SAM blocks. \textit{\enquote{rand init}} refers to SAM blocks with random initialization.
	}
	\label{tab:sam}
\end{table}

\begin{table}[t]
	\centering
	\resizebox{0.48\textwidth}{!}{%
		\begin{tabular}{cccc|cccccc}
			\toprule
			\multirow{2}{*}{Index} & \multicolumn{3}{c|}{Methods} & \multirow{2}{*}{A-847} & \multirow{2}{*}{PC-459} & \multirow{2}{*}{A-150} & \multirow{2}{*}{PC-59} & \multirow{2}{*}{PAS-20} & \multirow{2}{*}{PAS-20} \\
			& Point    & B-box    & Mask   & & & & & & \\
			\midrule\midrule
			(a) & \ding{51} & & & 15.2 & 23.6 & 37.8 & 59.4 & 96.7 & 81.3 \\
			(b) & & \ding{51} & & 14.3 & 20.5 & 33.5 & 59.4 & 96.5 & 81.4 \\
			(c) & \ding{51} & \ding{51} & & 15.1 & 23.4 & 38.1 & 62.0 & 96.7 & 82.5 \\
			(d) & \ding{51} & & \ding{51} & \textbf{18.1} & \textbf{27.0} & \underline{41.8} & \underline{65.6} & \textbf{98.3} & \textbf{86.3} \\
			(e) & & \ding{51} & \ding{51} & 17.0 & 26.1 & 39.9 & 62.0 & 97.1 & 84.2 \\
			(f) & \ding{51} & \ding{51} & \ding{51} & \underline{17.6} & \underline{26.5} & \textbf{41.9} & \textbf{65.8} & \underline{98.0} & \underline{86.1} \\      
			\bottomrule
		\end{tabular}
	}
	\caption{Quantitative performance comparison based on the combination of pseudo prompts used. \enquote{B-box} refers to a bounding box.}
	\label{tab:ppg}
\end{table}

\noindent
\textbf{Model efficiency.} In Table~\ref{tab:flops}, we compare the model efficiency of the proposed ESC-Net with previous methods~\cite{ding2022decoupling, xu2022simple, liang2023open, cho2024cat}. To this end, we calculate the model's parameters, inference time, and inference GFLOPs. All experiments were conducted on the same single RTX A6000 GPU, and except for the inference time, the rest of the data follows the results presented by~\cite{cho2024cat}. As shown in Table~\ref{tab:flops}, the proposed model exhibits lightweight and efficient inference performance compared to existing methods. Furthermore, it has comparable parameters and GFLOPs when compared to CAT-Seg, which is also a correlation-based one-stage model.

\begin{table}[t]
	\centering
	\resizebox{0.48\textwidth}{!}{%
		\begin{tabular}{c|ccc}
			\toprule
			Model     & \# of Params. (M) & Inference time (s) & Inference GFLOPs \\
			\midrule\midrule
			ZegFormer~\citep{ding2022decoupling} & 531.2 & 3.10 & 19,425.6         \\
			ZSSeg~\citep{xu2022simple}     & 530.8 & 3.11 & 22,302.1         \\
			OVSeg~\citep{liang2023open}     & 532.6  & 2.98 & 19,345.6         \\
			CAT-Seg~\citep{cho2024cat}   & 433.7 & 0.78 & 2,121.1          \\
			\midrule
			Ours      & 451.3 & 0.76 & 2,203.5          \\
			\bottomrule     
		\end{tabular}
	}
	\caption{Model efficiency comparison. All results are measured with a single RTX A6000 GPU.}
	\label{tab:flops}
\end{table}

\noindent
\textbf{Limitations.} While ESC-Net leverages SAM blocks for spatial aggregation, it operates on low-resolution feature levels, which limits the mask resolution and affects the precision of segmentation boundaries. This issue results in lower-quality masks compared to methods that can perform high-resolution aggregation. Additionally, similar to other correlation-based approaches, ESC-Net’s memory requirements increase with the number of classes due to the expanded text-image feature correlations. This growth in memory usage can reduce the efficiency of the model, particularly when handling a large variety of classes.

%% file: sec/5_conclusion.tex
\section{Conclusion}
In this paper, we present ESC-Net, a one-stage open-vocabulary semantic segmentation model that efficiently segments images across diverse class labels. By combining SAM’s powerful class-agnostic segmentation with CLIP’s vision-language alignment, ESC-Net overcomes the limitations of prior two-stage and one-stage methods. Our model uses correlation-based pseudo prompts from CLIP’s image-text features to guide SAM, generating precise prediction masks without requiring a separate mask proposal stage, thus reducing computational costs. We evaluate ESC-Net on standard open-vocabulary benchmarks, including ADE20K, PASCAL-VOC, and PASCAL-Context, where it achieves state-of-the-art performance. Additionally, ablation studies confirm its robustness across various challenging scenarios, demonstrating ESC-Net as an efficient and accurate solution for open-vocabulary semantic segmentation.